\documentclass[letterpaper, 10 pt, conference]{ieeeconf}  

\IEEEoverridecommandlockouts                              

\usepackage{graphics} 
\usepackage{epsfig} 
\usepackage{amsmath} 
\usepackage{amssymb}  
\usepackage{hyperref}
\usepackage{booktabs}
\usepackage{cite}

\let\labelindent\relax
\usepackage{enumitem}

\title{\LARGE \bf 3D Multi-Object Tracking: A Baseline and New Evaluation Metrics}

\author{Xinshuo Weng$^{1}$, Jianren Wang$^{1}$, David Held$^{1}$ and Kris Kitani$^{1} $
\thanks{$^{1}$Xinshuo Weng, Jianren Wang, David Held and Kris Kitani are with Robotics Institute, Carnegie Mellon University, USA. {\tt\small \{xinshuow, jianrenw, dheld, kkitani\}@cs.cmu.edu}.}
\vspace{-0.2cm}
}

\begin{document}

\maketitle
\thispagestyle{empty}
\pagestyle{empty}


\begin{abstract}

3D multi-object tracking (MOT) is an essential component for many applications such as autonomous driving and assistive robotics. Recent work on 3D MOT focuses on developing accurate systems giving less attention to practical considerations such as computational cost and system complexity. In contrast, this work proposes a simple real-time 3D MOT system. Our system first obtains 3D detections from a LiDAR point cloud. Then, a straightforward combination of a 3D Kalman filter and the Hungarian algorithm is used for state estimation and data association. Additionally, 3D MOT datasets such as KITTI evaluate MOT methods in the 2D space and standardized 3D MOT evaluation tools are missing for a fair comparison of 3D MOT methods. Therefore, we propose a new 3D MOT evaluation tool along with three new metrics to comprehensively evaluate 3D MOT methods. We show that, although our system employs a combination of classical MOT modules, we achieve state-of-the-art 3D MOT performance on two 3D MOT benchmarks (KITTI and nuScenes). Surprisingly, although our system does not use any 2D data as inputs, we achieve competitive performance on the KITTI 2D MOT leaderboard. Our proposed system runs at a rate of $207.4$ FPS on the KITTI dataset, achieving the fastest speed among all modern MOT systems. To encourage standardized 3D MOT evaluation, our system and evaluation code are made publicly available at \url{https://github.com/xinshuoweng/AB3DMOT}.

\end{abstract}


\section{Introduction}
\vspace{-0.1cm}

MOT is an essential component for many real-time applications such as autonomous driving \cite{Wang2018, Weng2020_spcsf} and assistive robotics \cite{Sun2020_imuvisual, Manglik2019}. Due to advancements in object detection~\cite{Ren2015, Shi2019, Lee2016, Weng20182}, there has been much progress on MOT. For example, for the car class on the KITTI~\cite{Geiger2012} 2D MOT benchmark, the MOTA (multi-object tracking accuracy) has improved from 57.03~\cite{Yoon2016} to 84.04~\cite{Karunasekera2019} in just two years! While we are encouraged by the progress, we observed that our focus on innovation and accuracy has come at the cost of practical factors such as computational efficiency and system simplicity. State-of-the-art methods typically require a large computational cost \cite{Sharma2018, Tian2019, Frossard2018, Baser2019} making real-time performance a challenge. Also, modern MOT systems are often very complex and it is not always clear which part of the system contributes the most to performance. For example, leading works~\cite{Baser2019, Frossard2018, Scheidegger2018} have substantially different system pipelines but only minor differences in performance. In these cases, modular comparative analysis is quite challenging. 

\begin{figure}[t]
\centering
\includegraphics[trim=0.3cm 2cm 9.5cm 0.3cm, clip=true, width=0.9\linewidth]{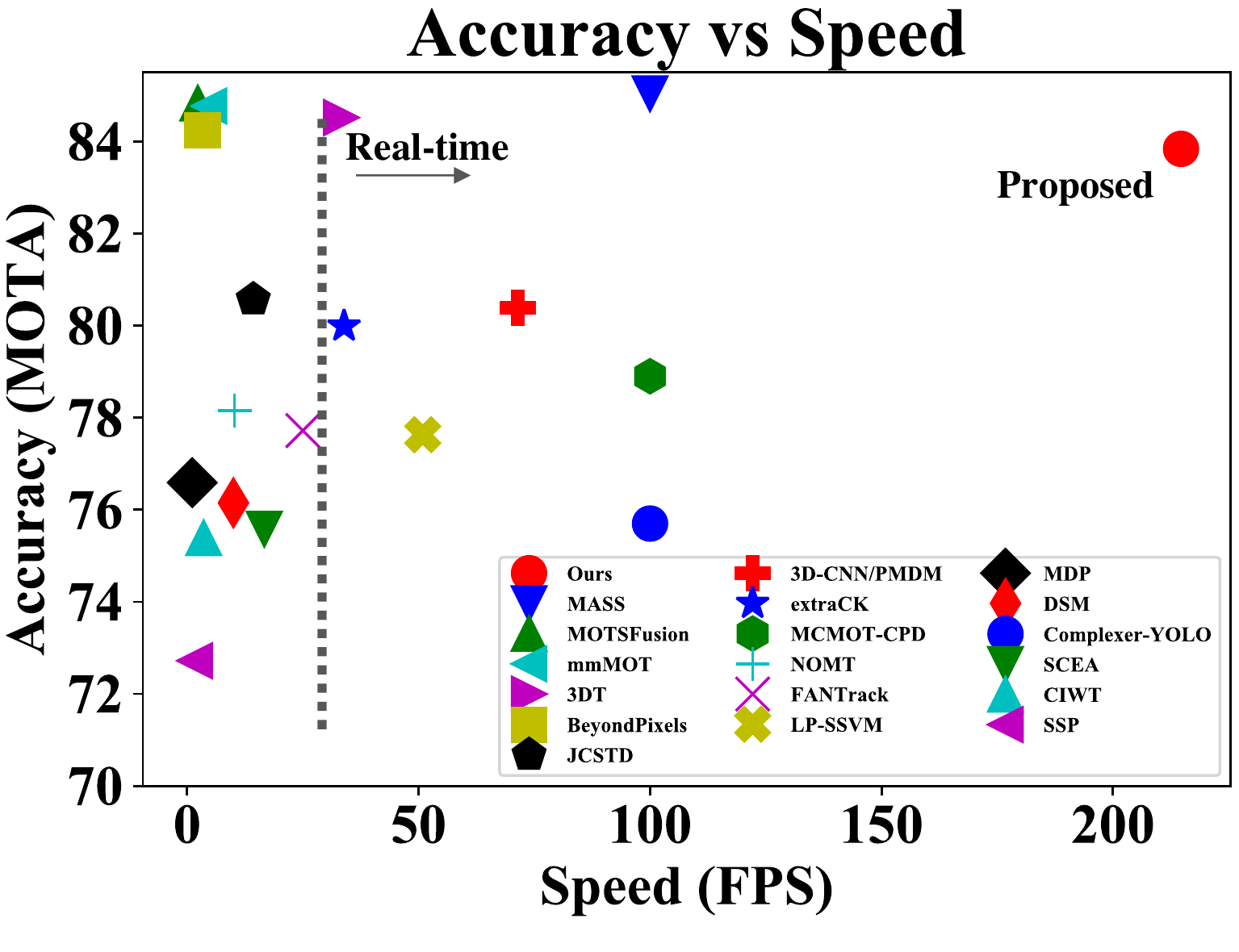}
\vspace{-0.6cm}
\caption{MOTA of modern 2D and 3D MOT systems on the KITTI 2D MOT leaderboard. The higher and more right is better. Our 3D MOT system achieves competitive MOTA in 2D MOT evaluation while being the fastest.}
\vspace{-0.5cm}
\label{fig:teaser}
\end{figure}

To provide a standard 3D MOT baseline for comparative analysis, we implement a classical approach which is both efficient and simple in design -- the Kalman filter \cite{Kalman1960} (1960) coupled with the Hungarian method \cite{WKuhn1955} (1955). Specifically, our system employs an off-the-shelf 3D object detector to obtain 3D detections from the LiDAR point cloud \cite{Shi2019}. Then, a combination of the 3D Kalman filter (with a constant velocity model) and the Hungarian algorithm is used for state estimation and data association. Unlike other filter-based MOT systems which define the state space of the filter in the 2D space \cite{Bewley2016} or bird's eye view \cite{Patil2019}, we extend the state space of the objects to the 3D space, including 3D location, 3D size, 3D velocity and heading orientation. 

Our empirical results are alarming. While the combination of modules in our system is straightforward, we achieve state-of-the-art 3D MOT performance on standard 3D MOT datasets: KITTI and nuScenes. Surprisingly, although our system does not use any 2D data as inputs, we also achieve competitive performance on the KITTI 2D MOT leaderboard as shown in Fig. \ref{fig:teaser}. We hypothesize that the strong 2D MOT performance of our 3D MOT system may be due to the fact that tracking in 3D can better resolve depth ambiguities and lead to fewer mismatches than tracking in 2D. Also, due to efficient design of our system, it runs at a rate of $207{.}4$ FPS on the KITTI dataset, achieving the fastest speed among modern MOT systems. To be clear, the contribution of this work is not to innovate 3D MOT algorithms but to provide a more clear picture of modern 3D MOT systems in comparison to a most basic yet strong baseline, the results of which are important to share across the community.

In addition to the 3D MOT system, we also observed two issues in 3D MOT evaluation: \textit{(1) Standard MOT benchmarks such as the KITTI dataset only supports 2D MOT evaluation}, \emph{i.e.}, evaluation on the image plane. A tool to evaluate 3D MOT systems in 3D space is not currently available. On the KITTI dataset, the convention to evaluate 3D MOT methods is to project the 3D MOT results to the image plane and then use the KITTI 2D MOT evaluation tool. However, we believe that this will hamper the future progress of 3D MOT systems as evaluation on the image plane cannot provide a fair comparison of 3D MOT methods, \emph{e.g.}, a system that achieves better tracking in 3D does not necessarily have higher performance in 2D MOT evaluation. To overcome the issue, we propose an MOT evaluation tool that evaluates MOT systems directly in 3D space using 3D metrics; \textit{(2) Common MOT metrics such as MOTA and MOTP do not consider the confidence score of tracked objects.} As a result, users must manually select a threshold and filter out tracked objects with lower scores. However, selecting the best threshold requires non-trivial efforts. Also, evaluation at a single threshold prevents us from understanding the full spectrum of accuracy and precision of a MOT system. To address the issue, we propose three new integral metrics to summarize the performance of MOT methods across many thresholds. We hope that our new evaluation tool including metrics will serve as a standard for future 3D MOT evaluation. Our contributions are summarized as follows: 
\begin{enumerate}
    \item We propose an accurate real-time 3D MOT system based on a 3D Kalman filter for online applications; 
    \item We propose a new 3D MOT evaluation tool along with three new metrics to standardize 3D MOT evaluation;
    \item Our 3D MOT system achieves S.O.T.A. performance and the fastest speed on standard 3D MOT datasets. 
\end{enumerate}


\begin{figure*}
\begin{center}
\includegraphics[trim=0.2cm 8.2cm 0cm 0cm, clip=true, width=\linewidth]{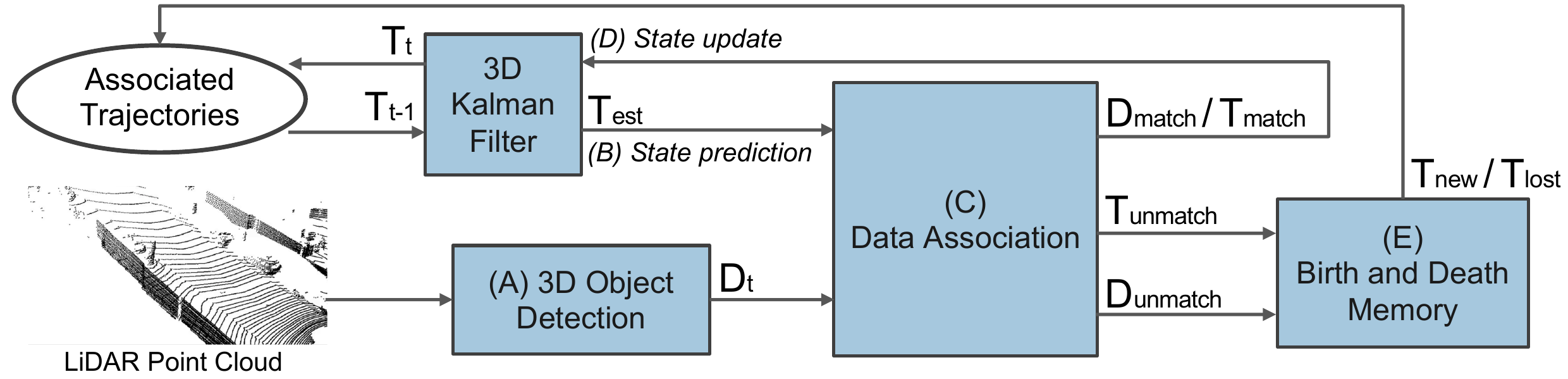}
\end{center}
\vspace{-0.6cm}
\caption{\textbf{Proposed System Pipeline}: (A) a 3D detection module obtains 3D detections $D_t$ from the LiDAR point cloud; (B) a 3D Kalman filter predicts the state of trajectories $T_{t-1}$ to the current frame $t$ as $T_\text{est}$ during the state prediction step; (C) the detections $D_t$ and predicted trajectories $T_\text{est}$ are associated using the Hungarian algorithm; (D) the state of each matched trajectory in $T_\text{match}$ is updated by the 3D Kalman filter based on the corresponding matched detection in $D_\text{match}$ to obtain the final trajectories $T_t$; (E) a birth and death memory takes the unmatched detections $D_\text{unmatch}$ and unmatched trajectories $T_\text{unmatch}$ as inputs and creates new trajectories $T_\text{new}$ and deletes disappeared trajectories $T_\text{lost}$ from the associated trajectories.}
\label{fig:pipeline}
\vspace{-0.5cm}
\end{figure*}


\section{Related Work}

\noindent\textbf{2D Multi-Object Tracking.} Recent 2D MOT systems can be split into batch and online methods based on data association. Batch methods attempt to find the global optimal association from the entire sequence. These methods often create a network flow graph and can be solved by the min-cost flow algorithms~\cite{Zhang2008, Schulter2017}. In contrast, online methods only require the information up to the current frame and are applicable for online applications. Online methods often formulate data association as a bipartite graph matching problem and solve it using the Hungarian algorithm \cite{WKuhn1955, Bewley2016}. Beyond using the Hungarian algorithm, modern online methods design deep association networks~\cite{Baser2019, Weng2020_gnn3dmot} that can construct the association using neural networks. Our proposed system falls into the category of online methods. For simple design and real-time efficiency, we do not use neural networks and only adopt the Hungarian algorithm.

To achieve data association, designing appropriate cost functions to measure similarity is crucial to a MOT system. Early work~\cite{Pirsiavash2015, Zhang2008} employs hand-crafted features such as spatial distance and color histograms as the cost function. Modern methods often use the motion model \cite{Choi2015, Bewley2016, Dicle2013} and the appearance feature \cite{Choi2015, Bae2014, yujhe2020}. For system simplicity, we only employ the simplest motion model, \emph{i.e.}, constant velocity, while not using any appearance cue.

\vspace{1.5mm}\noindent\textbf{3D Multi-Object Tracking.} 3D MOT systems often share the same components as 2D MOT systems. The distinction lies in that the input detections are in the 3D space instead of the image plane. Therefore, 3D MOT systems can obtain the motion and appearance information in the 3D space without perspective distortion. \cite{Scheidegger2018} proposed to estimate the distance of objects to the camera and their velocity in the 3D space as the motion cue. \cite{Patil2019} used an unscented Kalman filter to estimate the linear and angular velocity on the ground. \cite{Osep2017} proposed a 2D-3D Kalman filter to utilize the observation from the image and 3D world. Beyond using hand-crafted features, \cite{Zhang2019, Weng2020_gnn3dmot, Weng2020_gnntrkforecast, Wang2020_gnndettrk} used neural networks to learn the 3D appearance and motion features from data. Unlike prior work uses various 3D features and has complex systems, we only use a 3D Kalman filter to obtain the 3D motion cue for simplicity and efficiency, with extending the state space of the filter to full 3D domain including 3D location, 3D velocity, 3D size and heading orientation.


\vspace{-0.1cm}
\section{Approach}

The goal of 3D MOT is to associate 3D detections in a sequence. As our system is an online MOT system, at every timestamp, we only require detections in the current frame and associated trajectories from the previous frames. Our system pipeline is shown in Fig. \ref{fig:pipeline}: (A) a 3D detection module is used to obtain 3D detections from the LiDAR point cloud; (B) a 3D Kalman filter predicts the state of associated trajectories from the previous frames to the current frame; (C) a data association module matches the predicted trajectories from Kalman filter and detections in the current frame; (D) the 3D Kalman filter updates the state of matched trajectories based on the matched detections; (E) a birth and death memory creates trajectories for new objects and deletes trajectories for disappeared objects. Except for the pre-trained 3D detection module, our 3D MOT system does not need any training and can be directly used for inference.

\subsection{3D Object Detection}

Thanks to advancements in 3D object detection, we have access to high-quality detections. Here, we experiment with \cite{Shi2019, Weng2019} on KITTI and \cite{Zhu2019} on nuScenes. We directly use their pre-trained models on the corresponding dataset. In frame $t$, the output of 3D detection module is a set of detections $D_{t}=\{D^1_t, D_t^2, \cdots, D_t^{n_t}\}$ ($n_t$ is the number of detections). Each detection $D_t^j$, where $j\in \{1, 2, \cdots, n_t\}$, is represented as a tuple $(x, y, z, \theta, l, w, h, s)$, including location of the object center in the 3D space ($x$, $y$, $z$), object's 3D size ($l$, $w$, $h$), heading angle $\theta$ and confidence score $s$. We will show how different 3D detection modules affect the performance of our 3D MOT system in the experiments.

\subsection{3D Kalman Filter: State Prediction}

To predict the state of object trajectories from the previous frames to the current frame, we approximate objects' inter-frame displacement using a constant velocity model independent of camera ego-motion. That means we do not explicitly estimate the ego-motion but rely on our motion model to accommodate both the ego-motion and motion of the other objects. We formulate the state of an object trajectory as a 11-dimensional vector $T = (x, y, z, \theta, l, w, h, s, v_x, v_y, v_z)$, where the additional variables $v_x$, $v_y$, $v_z$ represent the object velocity in the 3D space. Note that we do not include the angular velocity $v_{\theta}$ in the state space for simplicity as we empirically found that including the angular velocity does not really improve the performance. In every frame, the state of associated trajectories from the previous frame $T_{t-1}$=$\{T_{t-1}^1, T_{t-1}^2, \cdots, T_{t-1}^{m_{t-1}}\}$ ($m_{t-1}$ is the number of trajectories in the frame $t$-$1$) will be propagated to the frame $t$ as $T_{\text{est}}$, based on the constant velocity model:
\vspace{-0.3cm}
\begin{equation}
    x_{\text{est}} = x + v_x,  \ \ \ \ \  y_{\text{est}} = y + v_y, \ \ \ \ \ z_{\text{est}} = z + v_z. \\
    \vspace{-0.2cm}
\end{equation}

As a result, for every trajectory $T_{t-1}^i$ in $T_{t-1}$ where $i$ $\in$ $\{$$1$, $2$, $\cdots$, $m_{t-1}$$\}$, the predicted state in the frame $t$ is $T_{\text{est}}^i$ = ($x_{\text{est}}$, $y_{\text{est}}$, $z_{\text{est}}$, $\theta$, $l$, $w$, $h$, $s$, $v_x$, $v_y$, $v_z$).

\subsection{Data Association}\label{sec:association}

To match the predicted trajectories $T_{\text{est}}$ with the detections $D_t$, we first construct the affinity matrix with a dimension of $m_{t-1}$$\times$$n_t$ by computing the 3D Intersection of Union (IoU) or negative center distance between every pair of the trajectory $T_{\text{est}}^i$ and detection $D_t^j$. Then, the data association becomes a bipartite graph matching problem, which can be solved in polynomial time using the Hungarian algorithm~\cite{WKuhn1955}. Also, we reject a matching if the 3D IoU is less than a threshold $\text{IoU}_\text{min}$ (or the center's distance is larger than a threshold $\text{dist}_\text{max}$ if using center distance to compute affinity matrix). The outputs of data association are as follows:
\begin{align}
\begin{split}\label{eq:type1}
    T_\text{match} ={}& \{T_\text{match}^1, T_\text{match}^2, \cdots, T_\text{match}^{w_t}\},
\end{split}\\
\begin{split}\label{eq:type2}
    D_\text{match} ={}& \{D_\text{match}^1, D_\text{match}^2, \cdots, D_\text{match}^{w_t}\},
\end{split}\\
\begin{split}\label{eq:type3}
    T_{\text{unmatch}} ={}& \{T_\text{unmatch}^1, T_\text{unmatch}^2, \cdots, T_\text{unmatch}^{m_{t-1}-w_t}\},
\end{split}\\
\begin{split}\label{eq:type4}
    D_\text{unmatch} ={}& \{D_\text{unmatch}^1, D_\text{unmatch}^2, \cdots, D_\text{unmatch}^{n_t-w_t}\},
\end{split}
\end{align}
where $T_\text{match}$ and $D_\text{match}$ are the matched trajectories and detections and $w_t$ denotes the number of matches. Also, $T_{\text{unmatch}}$ and $D_\text{unmatch}$ are the unmatched trajectories and detections. Note that, $T_\text{unmatch}$ is the complementary set of $T_{\text{match}}$ in $T_{\text{est}}$. Similarly, $D_\text{unmatch}$ is the complementary set of $D_{\text{match}}$ in $D_{\text{t}}$.

\subsection{3D Kalman Filter: State Update}\label{sec:update}

To account for the uncertainty of state prediction, we update the state of each trajectory in $T_\text{match}$ based on its corresponding detection in $D_\text{match}$. As a result, we obtain the final associated trajectories in frame $t$ as $T_t$=$\{T_t^1, T_t^2, \cdots, T_t^{w_t}\}$. Following the Bayes rule, the updated state of each trajectory $T_t^k$=$(x^\prime, y^\prime, z^\prime, \theta^\prime, l^\prime, w^\prime, h^\prime, s^\prime, v_x^\prime, v_y^\prime, v_z^\prime)$, where $k$ $\in$ $\{1, 2, \cdots, w_t\}$, is the weighted average between the state of $T_\text{match}^k$ and $D_\text{match}^k$. The weights are determined by the state uncertainty of the matched trajectory $T_\text{match}^k$ and detection $D_\text{match}^k$ (please refer to the Kalman filter~\cite{Kalman1960} for details).

Also, we observe that directly applying the Bayes update rule to orientation $\theta$ does not work well. For example, there might be the case where the orientation of detection $D_\text{match}^k$ is nearly opposite to the orientation of the corresponding trajectory $T_\text{match}^k$, \emph{i.e.}, differ by $\pi$. Although we know that this is impossible because objects should move smoothly and cannot change the orientation by $\pi$ in one frame (\emph{i.e.}, 0.1s in KITTI), the prediction of the orientation in either the detection or the trajectory can be wrong, making this scenario possible. As a result, if we follow the normal state update rule, the final trajectory $T_t^k$ in this case will have an orientation somewhere in the middle of the orientation of $D_\text{match}^k$ and $T_\text{match}^k$, which will lead to a low 3D IoU between the associated trajectory and the ground truth. To prevent this issue, we propose an orientation correction technique. When the difference of the orientation $\theta_d$ between $D_\text{match}^k$ and $T_\text{match}^k$ is greater than $\frac{\pi}{2}$, we add a $\pi$ to the orientation in $T_\text{match}^k$ so that $\theta_d$ is always less than $\frac{\pi}{2}$, \emph{i.e.,} the orientation $D_\text{match}^k$ and $T_\text{match}^k$ are roughly consistent without substantial change. 

\vspace{-0.1cm}
\subsection{Birth and Death Memory}

As tracked objects might leave the scene and new objects might enter the scene, a module to manage the birth and death of the objects is necessary. On one hand, we consider all unmatched detections $D_\text{unmatch}$ as potential new objects entering the scene. However, to avoid creating false positive trajectories, a new trajectory $T_\text{new}^p$ will not be created for the unmatched detection $D_\text{unmatch}^p$ until $D_\text{unmatch}^p$ has been continually matched in the next $\text{Bir}_\text{min}$ frames, where $p\in \{1, 2, \cdots, n_t-w_t\}$. Once the new trajectory $T_\text{new}^p$ is created, we initialize its state same as its most recent detection $D_\text{unmatch}^p$ with zero velocity for $v_x$, $v_y$ and $v_z$. 

On the other hand, we consider all unmatched trajectories $T_\text{unmatch}$ as potential objects leaving the scene. However, to prevent deleting true positive trajectories that still exist in the scene but cannot find a match due to missing detection, we keep tracking each unmatched trajectory $T_\text{unmatch}^q$ for $\text{Age}_{\text{max}}$ frames before ensuring $T_\text{unmatch}^q$ is a disappeared trajectory $T_\text{lost}^q$, where $q\in \{1, 2, \cdots, m_{t-1}-w_t\}$, and deleting it from the set of associated trajectories. Ideally, true positive trajectories with missing detection can be interpolated by our 3D MOT system without being deleted, and only the trajectories that leave the scene are deleted.

\begin{figure*}
\begin{center}
\begin{minipage}[c]{0.245\textwidth}
    \begin{center}
    \includegraphics[width=\linewidth]{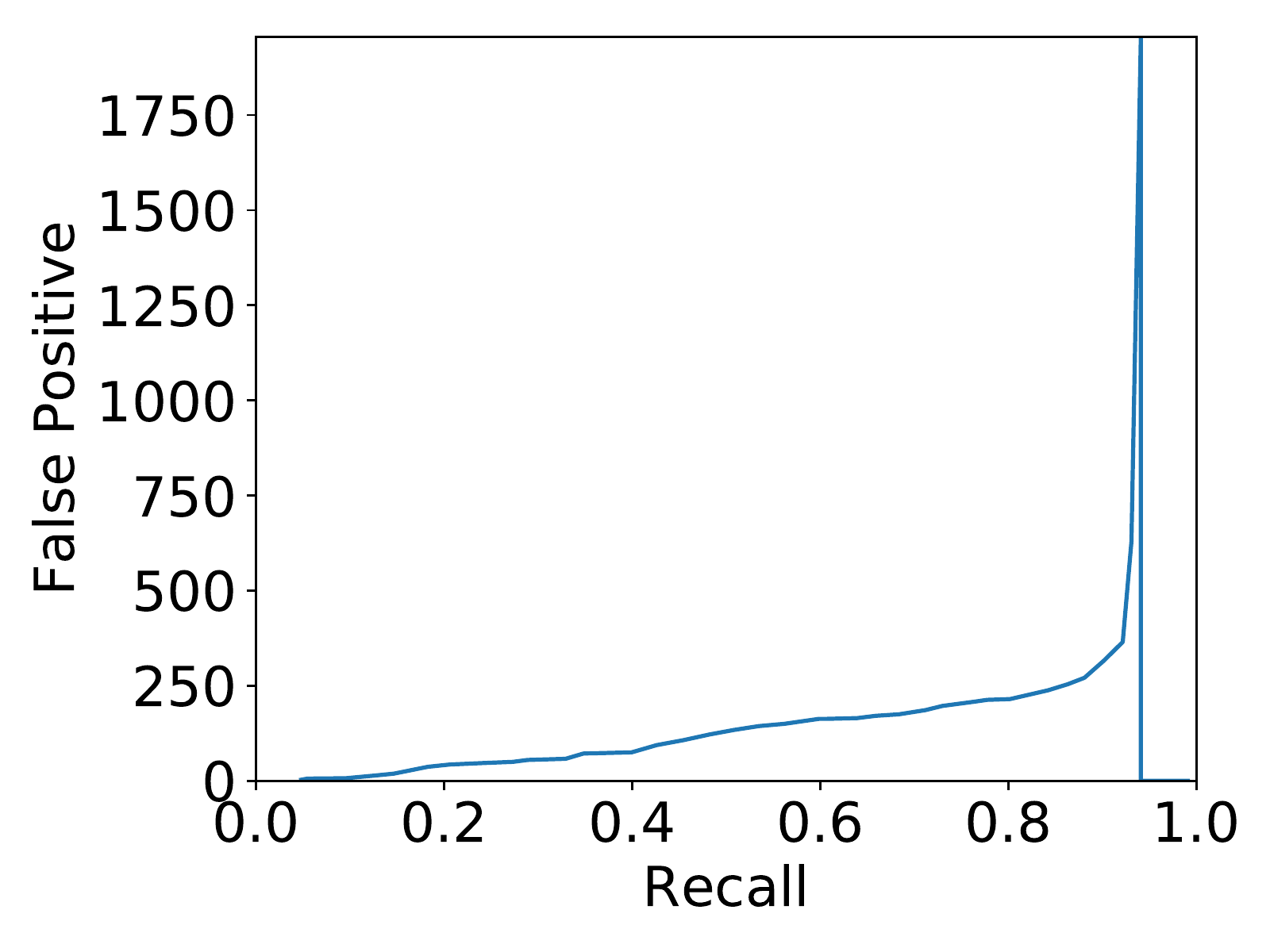}
    \vspace{-0.7cm}
    \\ (a) FP - Recall Curve
    \label{fig:fp_recall} 
    \end{center}
\end{minipage}
\begin{minipage}[c]{0.245\textwidth}
    \begin{center}
    \includegraphics[width=\linewidth]{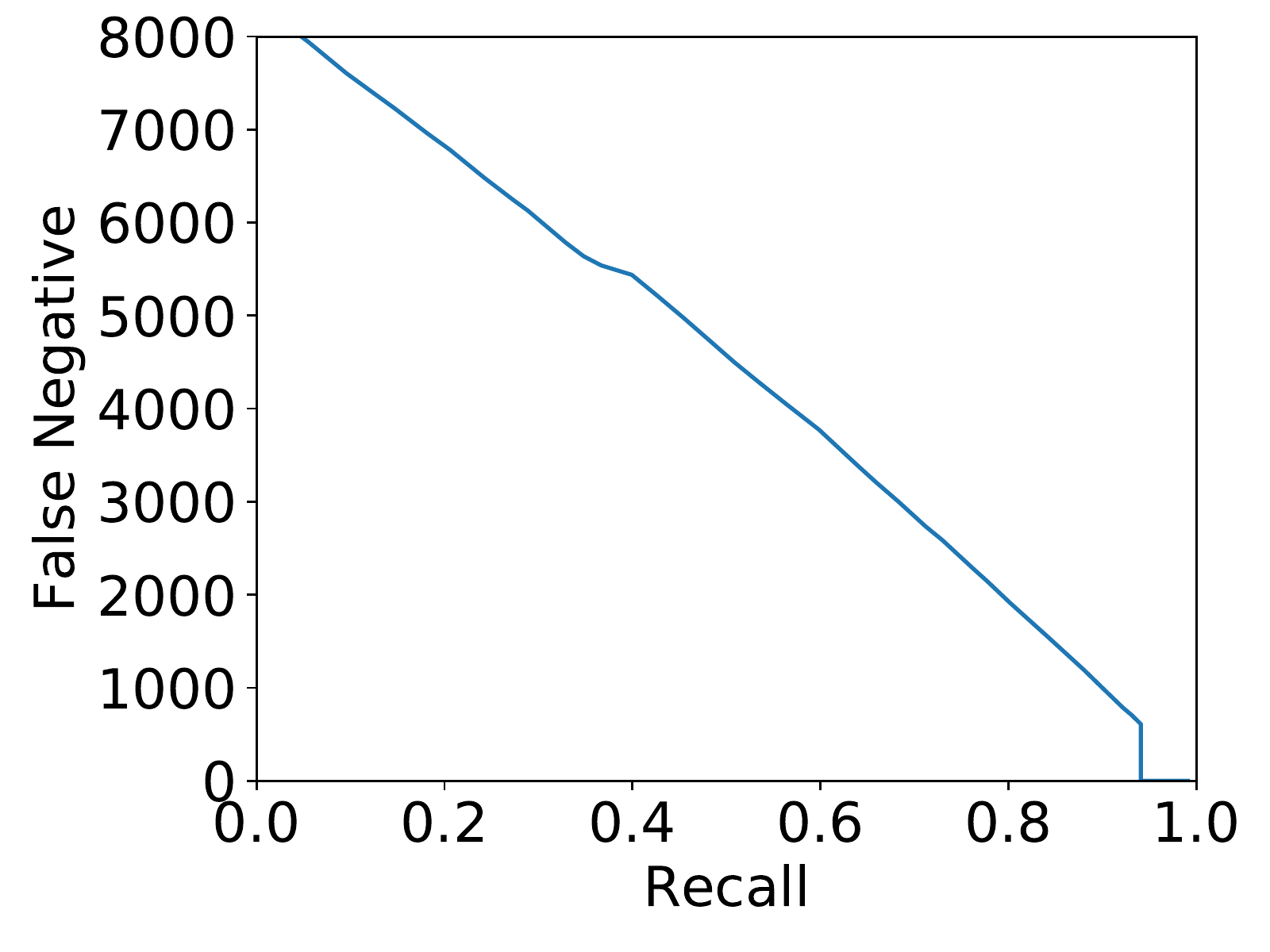}
    \vspace{-0.7cm}
    \\ (b) FN - Recall Curve
    \label{fig:fn_recall} 
    \end{center}
\end{minipage}
\begin{minipage}[c]{0.245\textwidth}
    \begin{center}
    \centering
    \includegraphics[width=\linewidth]{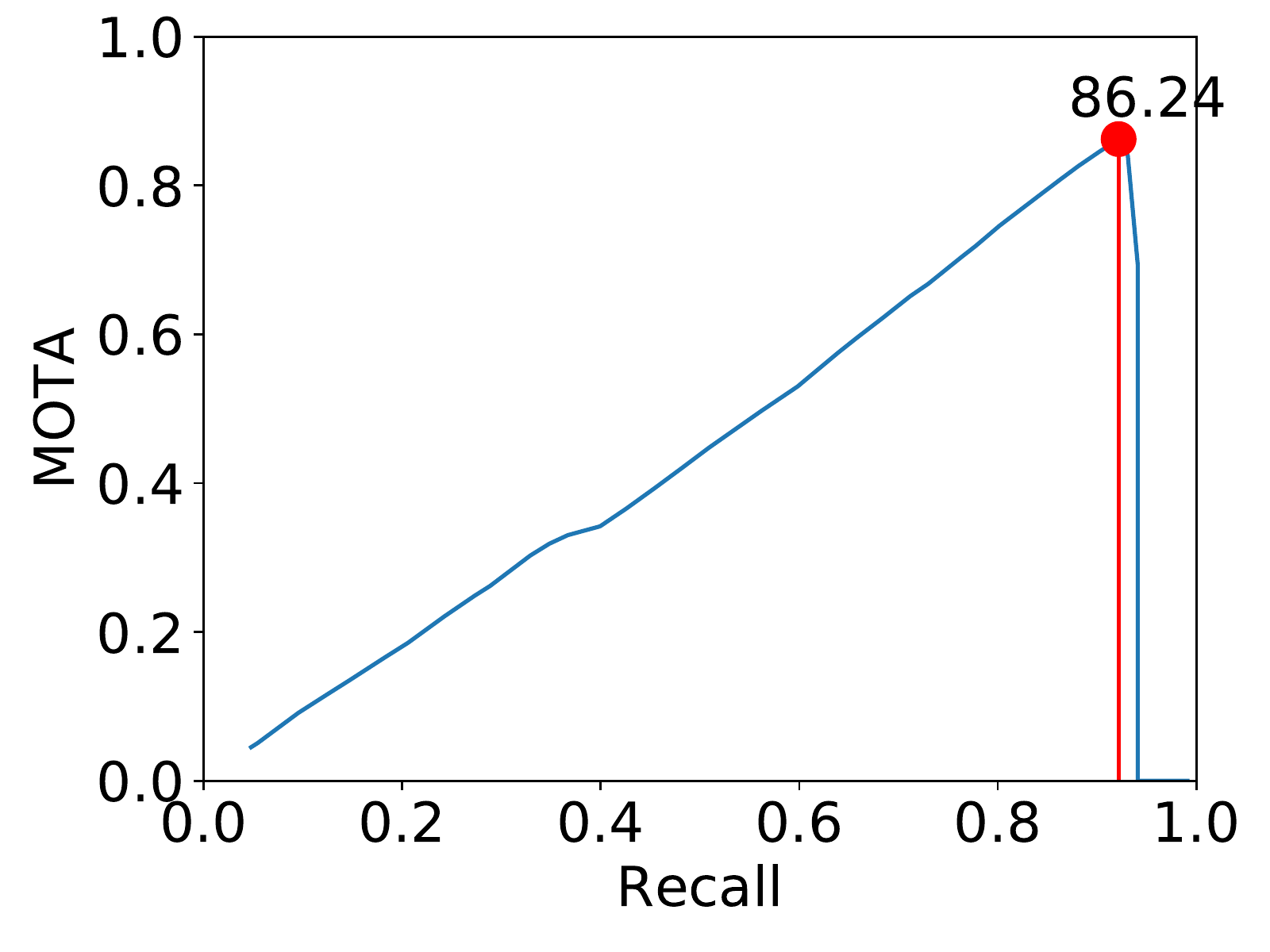}
    \vspace{-0.7cm}
    \\ (c) MOTA - Recall Curve
    \label{fig:mota_recall} 
    \end{center}
\end{minipage}
\begin{minipage}[c]{0.2455\textwidth}
    \begin{center}
\includegraphics[trim=0cm 0cm 0cm 0cm, clip=true, width=\linewidth]{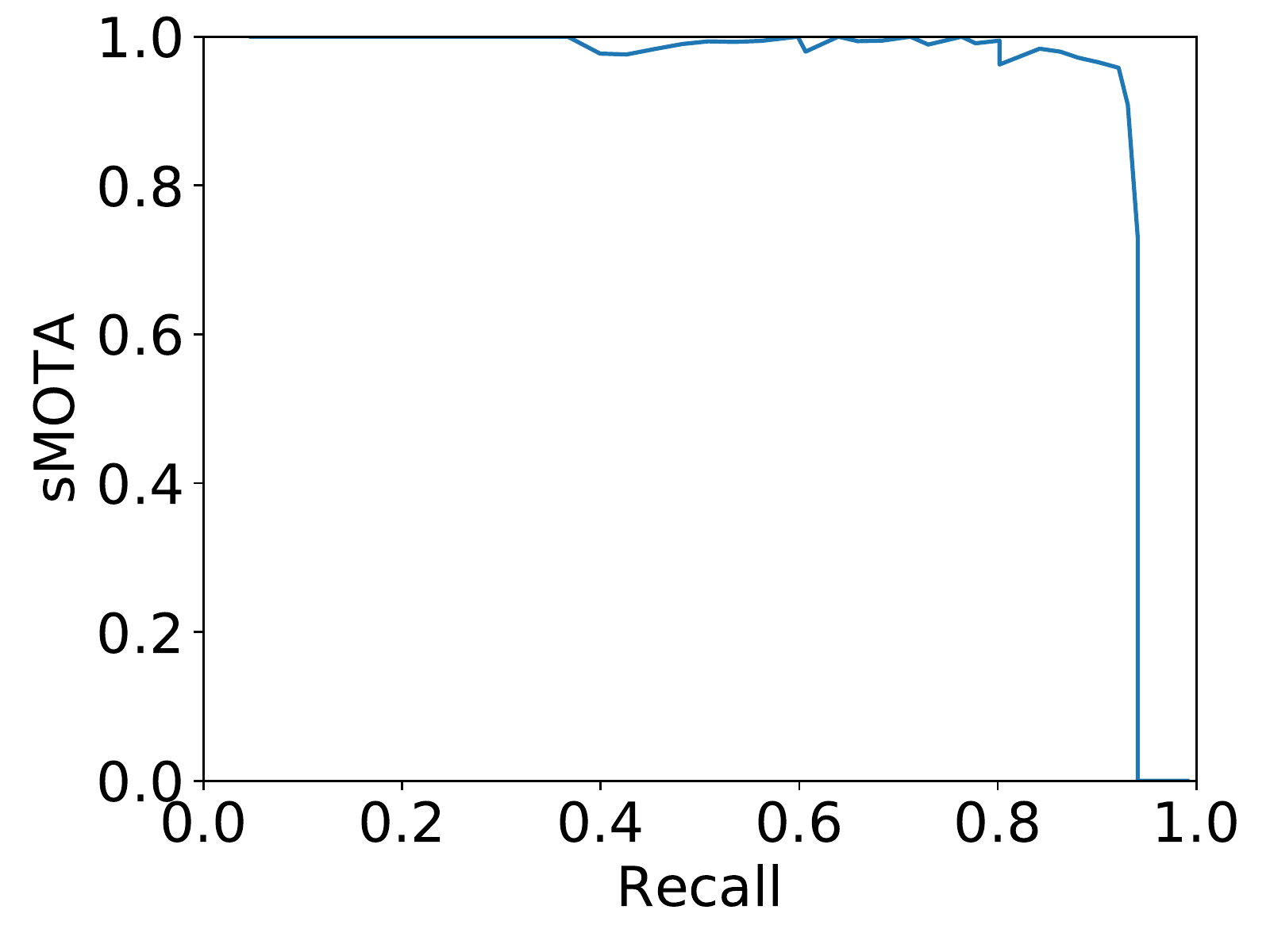}
    \vspace{-0.7cm}
    \\ (d) sMOTA - Recall Curve
    \label{fig:smota_recall} 
    \end{center}
\end{minipage}
\end{center}
\vspace{-0.35cm}
\caption{\textbf{(a)(b)(c) The effect of confidence threshold} on the CLEAR metrics: MOTA, FN and FP. We evaluate our 3D MOT system on the KITTI dataset using the proposed 3D MOT evaluation tool. We show that, to achieve the highest MOTA, a proper confidence threshold needs to be selected, otherwise the performance of MOTA will be decreased significantly due to a large number of false positives or false negatives. \textbf{(d) Effect of scale adjustment in MOTA}: the proposed scaled accuracy sMOTA has an upper bounding of $100\%$ at any recall value.}
\label{fig:evaluation}
\vspace{-0.5cm}
\end{figure*}


\vspace{-0.1cm}
\section{New 3D MOT Evaluation Tool}

As the pioneering 3D MOT benchmark, KITTI \cite{Geiger2012} dataset is crucial to the progress of 3D MOT systems. Though the KITTI dataset provides 3D object trajectories but it only supports 2D MOT evaluation, \emph{i.e.}, evaluation on the image plane, and a tool to evaluate 3D MOT systems directly in 3D space is not currently available. On the KITTI dataset, the current convention of evaluating 3D MOT systems is to project the 3D tracking results to the image plane and then use the KITTI 2D MOT evaluation tool, which matches the projected tracking results with ground truth trajectories on the image plane using 2D IoU as the cost function. However, we believe this will hamper the future progress of 3D MOT systems as evaluating on the image plane cannot provide a fair comparison of 3D MOT systems. For example, a system that outputs 3D trajectories with wrong depth estimates and low 3D IoU with the ground truth can still obtain high performance in 2D MOT evaluation as long as the projection of 3D trajectory outputs on the image plane has high 2D IoU with the ground truth on the image plane. 

To provide a fair comparison of 3D MOT systems, we implement an extension to the KITTI 2D MOT evaluation tool for 3D MOT evaluation. Specifically, we modify the cost function from 2D IoU to 3D IoU and match the 3D tracking results with 3D ground truth trajectories directly in 3D space. In this way, we no longer need to project our 3D tracking results to the image plane for evaluation. For every tracked object, its 3D IoU with ground truth is required to be above a threshold ${\text{IoU}}_\text{thres}$ (or center distance must be below a threshold ${\text{Dist}}_\text{thres}$) in order to be considered as a successful match. Although the extension of our 3D MOT evaluation tool is straightforward, we hope that it can serve as a standard to evaluate future 3D MOT systems.


\vspace{-0.1cm}
\section{New MOT Evaluation Metrics}
\subsection{Limitation of the CLEAR Metrics}

Conventional MOT evaluation is based on CLEAR metrics \cite{Bernardin2008} such as MOTA (see Section \ref{sec:metrics} for details), MOTP, FP, FN, Precision, F1 score, IDS, FRAG. However, none of these metrics explicitly consider the object's confidence score $s$. In other words, the CLEAR metrics consider all object trajectories having the same confidence $s$=$1$, which is an unreasonable assumption because there could be many false positive trajectories with low confidence scores. Therefore, to reduce the number of false positives and achieve a high MOTA\footnote{MOTA is the primary metric for ranking in most MOT benchmarks.}, users must manually select a threshold and filter out tracked objects with confidence scores lower than the threshold prior to submitting the results for evaluation.

Our observations to the above evaluation are two-fold: (1) selecting the best threshold for a 3D MOT system requires non-trivial efforts from the users and the confidence threshold can be significantly different if a 3D MOT system changes its input detections or is being evaluated on a different dataset. As a result, users must run extensive experiments on the validation set to tune the confidence threshold; (2) using a single confidence threshold for evaluation prevents us from understanding how the performance of 3D MOT systems changes as a function of the threshold. In fact, we observed that different confidence thresholds can significantly affect the performance of the CLEAR metrics. For example, we show the performance of our system on three metrics at different thresholds in Fig. \ref{fig:evaluation} using the data from the car subset of the KITTI MOT dataset. To generate the results, we first sort the tracking results based on the confidence score $s$\footnote{We define the confidence score of an object trajectory as the average of its confidence scores across all frames.}. Then, we define a set of confidence thresholds based on the recall of our system between $0$ to $1$ with an interval of $0.025$. This results in 40 confidence thresholds excluding the confidence threshold which corresponds to the recall of $0$. For each confidence threshold, we evaluate the results using only trajectories with confidence higher than the threshold. We show that, in Fig. \ref{fig:evaluation} (a), the confidence threshold should not be very small (recall not very high) because the number of false positives will increase drastically, especially when the recall reaches $0.95$. Also, in Fig. \ref{fig:evaluation} (b), the confidence threshold should not be very large, \emph{i.e.}, recall should not be very small, as it results in a large number of false negatives. As a result, in Fig. \ref{fig:evaluation} (c), we observed that the highest MOTA value is achieved only when we choose a confidence threshold corresponding to the recall of $0.9$ which balances the false positives and false negatives.

Based on the above observations, we believe that using a single confidence threshold for evaluation requires non-trivial efforts from users, and more importantly, prevents us from understanding full spectrum of accuracy of a MOT system. One consequence is that a MOT system with a high MOTA at a single threshold and low MOTA at other thresholds can be still ranked high on the leaderboard. But ideally, we should aim to develop MOT systems that achieve high MOTA across many thresholds, \emph{i.e.}, 3D MOT systems that achieve high performance when using different detections as inputs. Prior work \cite{Solera2015} shares the same spirit with us in that \cite{Solera2015} also believes it is important to understand the performance of MOT systems at many operating points. Specifically, \cite{Solera2015} computes a MOTA matrix at different recall and precision values, similar to our MOTA-over-recall curve. The distinction lies in that, we additionally propose integral metrics (see Sec. \ref{sec:newmetrics}) that summarize the performance at many operating points into a single scalar for easy comparison.

\vspace{-0.1cm}
\subsection{Integral Metrics: AMOTA and AMOTP}\label{sec:newmetrics}

To deal with the issue that current MOT evaluation metrics do not consider the confidence and only evaluate at a single threshold, we propose two integral metrics -- AMOTA and AMOTP (average MOTA and MOTP) -- to summarize the performance of MOTA and MOTP across many thresholds. The AMOTA and AMOTP are computed by integrating MOTA and MOTP values over all recall values, \emph{e.g.,} area under the MOTA over recall curve for computing AMOTA. Similar to other integral metrics such as the average precision used in object detection, we approximate the integration with a summation over a discrete set of recall values. Specifically, given the original definition of the MOTA metric from \cite{Bernardin2008}:
\vspace{-0.15cm}
\begin{equation}
\text{MOTA} = 1 - \frac{\text{FP} + \text{FN} + \text{IDS}}{\text{num}_{\text{gt}}},
\vspace{-0.15cm}
\end{equation}
where $\text{num}_{\text{gt}}$ is the number of ground truth objects in all frames. The AMOTA is then defined as follows:
\vspace{-0.1cm}
\begin{equation}
\text{AMOTA} = \frac{1}{L} \sum_{r \in \{\frac{1}{L}, \frac{2}{L}, \cdots, 1\}} (1 - \frac{\text{FP}_r + \text{FN}_r + \text{IDS}_r}{\text{num}_{\text{gt}}}),
\vspace{-0.1cm}
\end{equation}
where $\text{FP}_r$, $\text{FN}_r$ and $\text{IDS}_r$ are the number of false positives, false negatives and identity switches computed at a specific recall value $r$. Also, $L$ is the number of recall values (number of confidence thresholds for integration). The higher $L$ is, more accurate the approximate integration can be. However, a large $L$ requires significant compute during evaluation. To balance the accuracy and speed, we use $40$ recall values (\emph{i.e.}, from $0\%$ to $100\%$ with an interval of $2.5\%$ excluding $0\%$), \emph{i.e.}, $L$=$40$. For a 3D MOT system which has a maximum recall of $r_m$ less than $100\%$, the MOTA values for integration beyond $r_m$ are $0$. As a result, our proposed metrics are biased towards high-recall systems. We believe that this bias is acceptable as having a high recall is crucial to prevent collision for autonomous systems in practice. Note that our proposed AMOTA metric is similar to the PR-MOTA metric proposed in the independent work \cite{Wen2020}.

\subsection{Scaled Accuracy Metric: sAMOTA}

Conventionally, an integral metric such as average precision is a percentage ranging from $0\%$ to $100\%$ so that it is easy to measure the absolute performance of the system. To ensure that the integral metric has a range between $0\%$ and $100\%$, the metric used at every operating point to compute the integral metric should also be between $0\%$ and $100\%$. However, we observe in Fig. \ref{fig:evaluation} (c) that the MOTA is likely to have a strict upper bound lower than $100\%$ at many recall values. In fact, the upper bound of the MOTA at a specific recall value $r$ is derived as follows:
\begin{align}
\begin{split}\label{eq:5}
\begin{aligned}
    \text{MOTA}_r &= 1 - \frac{\text{FP}_r + \text{FN}_r + \text{IDS}_r}{\text{num}_{\text{gt}}} \leq 1 - \frac{\text{FN}_r}{\text{num}_{\text{gt}}} \\ 
    &\leq 1 - \frac{\text{num}_{\text{gt}} \times (1 - r)}{\text{num}_{\text{gt}}} = r.
\end{aligned}
\end{split}
\end{align}
\vspace{-0.35cm}

\noindent The first inequality is true because the false positives $\text{FP}_r$ and identity switches $\text{IDS}_r$ are always non-negative. Also, the second inequality uses the fact that $\text{FN}_r \geq \text{num}_{\text{gt}} \times (1 - r)$ because if the recall is $r$ that means that at least $(1-r)$ of the total objects ($\text{num}_{\text{gt}}$) are not tracked. If $r$ is the upper bound on MOTA$_r$ then it follows that the integral metric AMOTA is upper bounded by $50\%$ (\emph{i.e.}, upper bound $r$ creates a triangle in the MOTA vs Recall Curve).

To make the value of the integral metric AMOTA range from $0\%$ to $100\%$, we need to scale the range of the $\text{MOTA}_r$. From Eq. \ref{eq:5}, we find that the reason why the $\text{MOTA}_r$ has a strict upper bound of $r$ is due to the fact that $\text{FN}_r \geq \text{num}_{\text{gt}} \times (1 - r)$. To adjust the $\text{MOTA}_r$, we propose two new metrics, called sMOTA (\textbf{s}caled MOTA) and sAMOTA (\textbf{s}caled AMOTA), which are defined as follows:
\vspace{-0.1cm}
\begin{equation}
\resizebox{0.91\hsize}{!}{\label{eq:smota}
    $\text{sMOTA}_r = \max(0 , 1 - \frac{\text{FP}_r + \text{FN}_r + \text{IDS}_r - (1 - r) \times \text{num}_{\text{gt}} }{r \times \text{num}_{\text{gt}}})$,
}
\vspace{-0.25cm}
\end{equation}
\vspace{-0.2cm}
\begin{equation}
    \text{sAMOTA} = \frac{1}{L} \sum_{r \in \{\frac{1}{L}, \frac{2}{L}, \cdots, 1\}} \text{sMOTA}_r,
\vspace{-0.1cm}
\end{equation}
\noindent with the number of objects $\text{num}_{\text{gt}} \times (1 - r)$ being subtracted from the $\text{FN}_r$ in the numerator, the proposed $\text{sMOTA}_r$ is now upper bounded by $100\%$, leading to that the $\text{sAMOTA}$ is upper bounded by $100\%$ as well. Note that we also add a scalar factor $r$ in the denominator as we think using the actual number of ground truth objects available at a recall value of $r$ (\emph{i.e.}, $r \times \text{num}_{\text{gt}}$) makes more sense than using the total number of objects $\text{num}_{\text{gt}}$, some of which are not even available to be tracked at a recall of $r$. Additionally, we add a max operation over zero in Eq. \ref{eq:smota}, which is to adjust the lower bound of the $\text{sMOTA}_r$ to zero. Otherwise, $\text{sMOTA}_r$ can approach towards negative if there are many false positives or identity switches. As a result, the proposed $\text{sMOTA}_r$ in Eq. \ref{eq:smota} can have a range between $0\%$ and $100\%$ as shown in Fig. \ref{fig:evaluation} (d), which also leads to the corresponding integral metric $\text{sAMOTA}$ having a range between $0\%$ and $100\%$. In summary, we believe that the proposed new integral metric -- $\text{sAMOTA}$, $\text{AMOTA}$, $\text{AMOTP}$ -- are able to summarize performance of MOT systems across all thresholds.


\begin{table*}[t]
\caption{Performance of car on the KITTI val set using the proposed \textbf{3D} MOT evaluation tool with new metrics.}
\vspace{-0.3cm}
\centering
\resizebox{\textwidth}{!}{
\begin{tabular}{@{}llrrrrrrrrrr@{}}
\toprule
Method & Input Data & Matching criteria & \textbf{sAMOTA}$\uparrow$ & AMOTA$\uparrow$ & AMOTP$\uparrow$ & MOTA$\uparrow$ & MOTP$\uparrow$ & IDS$\downarrow$ & FRAG$\downarrow$ & FPS$\uparrow$ \\
\midrule
mmMOT~\cite{Zhang2019} (ICCV$'$19) & 2D + 3D 
 & $\text{IoU}_{\text{thres}}$ = 0.25 & 70.61 & 33.08 & 72.45 & 74.07 & 78.16 & 10 & 55 & 4.8 (GPU) \\
& & $\text{IoU}_{\text{thres}}$ = 0.5 & 69.14 & 32.81 & 72.22 & 73.53 & 78.51 & 10 & 64 \\ 
& & $\text{IoU}_{\text{thres}}$ = 0.7 & 63.91 & 24.91 & 67.32 & 51.91 & 80.71 & 24 & 141 \\ 
                                        
FANTrack~\cite{Baser2019} (IV$'$20) & 2D + 3D 
 & $\text{IoU}_{\text{thres}}$ = 0.25 & 82.97 & 40.03 & 75.01 & 74.30 & 75.24 & 35 & 202 & 25.0 (GPU) \\
& & $\text{IoU}_{\text{thres}}$ = 0.5 & 80.14 & 38.16 & 73.62 & 72.71 & 74.91 & 36 & 211 \\
& & $\text{IoU}_{\text{thres}}$ = 0.7 & 62.72 & 24.71 & 66.06 & 49.19 & 79.01 & 38 & 406 \\
\midrule
\textbf{Ours} & 3D 
 & $\text{IoU}_{\text{thres}}$ = 0.25 & 93.28 & 45.43 & 77.41 & 86.24 & 78.43 & 0 & 15 & 207.4 (CPU) \\
& & $\text{IoU}_{\text{thres}}$ = 0.5 & 90.38 & 42.79 & 75.65 & 84.02 & 78.97 & 0 & 51 & \\
& & $\text{IoU}_{\text{thres}}$ = 0.7 & 69.81 & 27.26 & 67.00 & 57.06 & 82.43 & 0 & 157 &  \\
\bottomrule
\end{tabular}}
\vspace{-0.5cm}
\label{tab:3dcomparison}
\end{table*}

\begin{table}[t]
\caption{Performance of pedestrian and cyclist on KITTI val set.}
\vspace{-0.2cm}
\centering
\resizebox{\hsize}{!}{
\begin{tabular}{@{}lrrrrr@{}}
\toprule
Category & Matching criteria & \textbf{sAMOTA}$\uparrow$ & AMOTA$\uparrow$ & AMOTP$\uparrow$ & MOTA$\uparrow$ \\
\midrule
Pedestrian & $\text{IoU}_{\text{thres}}$ = 0.25 & 75.85 & 31.04 & 55.53 & 70.90 \\
            & $\text{IoU}_{\text{thres}}$ = 0.5 & 70.95 & 27.31 & 52.45 & 65.06 \\
Cyclist    & $\text{IoU}_{\text{thres}}$ = 0.25 & 91.36 & 44.34 & 79.18 & 84.87 \\
            & $\text{IoU}_{\text{thres}}$ = 0.5 & 89.27 & 42.39 & 77.56 & 79.82 \\
\bottomrule
\end{tabular}}
\vspace{-0.3cm}
\label{tab:3dcomparison_pedcyc}
\end{table}

\begin{table}[t]
\caption{Performance over all categories on the nuScenes val set.}
\vspace{-0.25cm}
\centering
\resizebox{\hsize}{!}{
\begin{tabular}{@{}lrrrrr@{}}
\toprule
Method       & Matching criteria & \textbf{sAMOTA}$\uparrow$ & AMOTA$\uparrow$ & AMOTP$\uparrow$ & MOTA$\uparrow$ \\
\midrule
FANTrack~\cite{Baser2019} & $\text{Dist}_{\text{thres}}$ = 2 & 19.64 & 2.36 & 22.92 & 18.60 \\
mmMOT~\cite{Zhang2019}    & $\text{Dist}_{\text{thres}}$ = 2 & 23.93 & 2.11 & 21.28 & 19.82 \\ 
\midrule
\textbf{Ours}             & $\text{Dist}_{\text{thres}}$ = 2 & 39.90 & 8.94 & 29.67 & 31.40 \\
\bottomrule
\end{tabular}}
\vspace{-0.7cm}
\label{tab:3dcomparison_nuscene}
\end{table}


\vspace{-0.1cm}
\section{Experiment}
\subsection{Settings}

\noindent\textbf{Evaluation Metrics.}\label{sec:metrics} In addition to the proposed sAMOTA, AMOTA and AMOTP, we also evaluate on standard CLEAR metrics such as MOTA, MOTP (multi-object tracking precision), IDS (number of identity switches), FRAG (number of trajectory fragmentation), FPS (frame per second).

\vspace{1.5mm}\noindent\textbf{Datasets.} We evaluate on the KITTI and nuScenes 3D MOT datasets, which provide LiDAR point cloud and 3D bounding box trajectories. As the KITTI test set only supports 2D MOT evaluation and its ground truth is not released to users, we have to use the KITTI val set for 3D MOT evaluation. Also, we are collaborating with nuTomony to use our proposed metrics to build 3D MOT evaluation on the nuScenes dataset. However, the first nuScenes 3D MOT challenge is not yet finished when this work was developed. As such, we use our evaluation tool to evaluate 3D MOT systems on the nuScenes val set to develop a temporary comparison. For future evaluation on the nuScenes dataset, we recommend users to use the evaluation code provided by nuScenes and primarily evaluate 3D MOT systems on the nuScenes test set for comparison, though our developed temporary comparison on the val set can still be used for reference. 

In terms of the data split, we follow \cite{Scheidegger2018} on KITTI and use sequences 1, 6, 8, 10, 12, 13, 14, 15, 16, 18, 19 as the val set and other sequences as the train set, through our 3D MOT system does not require training. For nuScenes, we use its default data split. Regarding object category, we follow the KITTI convention and show results on each category (Car, Pedestrian, Cyclist). For nuScenes, we first obtain results on each category and then compute the final performance by averaging over $7$ categories (Car, Truck, Trailer, Pedestrian, Bicycle, Motorcycle, Bus). For matching criteria, we follow the convention in KITTI 3D object detection benchmark and use 3D IoU to determine a successful match. Specifically, we use 3D IoU threshold $\text{IoU}_{\text{thres}}$ of 0.25, 0.5 for Pedestrian and Cyclist, and $\text{IoU}_{\text{thres}}$ of 0.25, 0.5, 0.7 for Car. On nuScenes, we follow the criteria defined in the nuScenes challenge and use a center distance $\text{Dist}_{\text{thres}}$ of $2$ meters.

\vspace{1.5mm}\noindent\textbf{Baselines.} We compare against modern open-sourced 3D MOT systems such as FANTrack \cite{Baser2019} and mmMOT \cite{Zhang2019}. We use the same 3D detections obtained by PointRCNN \cite{Shi2019} on KITTI and by Megvii \cite{Zhu2019} on nuScenes for our proposed method and baselines \cite{Baser2019,Zhang2019} that require 3D detections as inputs. For baseline \cite{Baser2019} that also requires the 2D detections as inputs, we use the 2D projection of 3D detections. 

\vspace{1.5mm}\noindent\textbf{Implementation Details.} For our best results in Table \ref{tab:3dcomparison}, \ref{tab:3dcomparison_nuscene}, \ref{tab:3dcomparison_pedcyc} and \ref{tab:ablation}, we use ($x$, $y$, $z$, $\theta$, $l$, $w$, $h$, $s$, $v_x$, $v_y$, $v_z$) as the state space of our 3D Kalman filter without including the angular velocity $v_{\theta}$. We use $F_\text{min}$=$3$ and $\text{Age}_\text{min}$=$2$ in the birth and death memory module. For the threshold to reject a matching in the data association module, we empirically found that using $\text{IoU}_{\text{min}}$=$0.01$ for Car, $\text{Dist}_{\text{max}}$=$1$ for Pedestrian, $\text{Dist}_{\text{max}}$=$6$ for Cyclist can obtain the best performance on the KITTI dataset. On the nuScenes dataset, we use $\text{Dist}_{\text{max}}$=$10$ for all object categories. For other detailed hyper-parameters, please directly check our code.

\begin{table*}[t]
\centering
\caption{Ablation study for Car on the KITTI val set using the proposed 3D MOT evaluation tool with new metrics.}
\vspace{-0.2cm}
\resizebox{\textwidth}{!}{
\begin{tabular}{@{}lrrrrrrrrrr@{}}
\toprule
Method variants & Matching criteria & \textbf{sAMOTA}$\uparrow$ & AMOTA$\uparrow$ & AMOTP$\uparrow$ & MOTA$\uparrow$ & MOTP$\uparrow$ & IDS$\downarrow$ & FRAG$\downarrow$ & FP$\downarrow$ & FN$\downarrow$ \\
\midrule
(a) replace detector with \cite{Weng2019} & $\text{IoU}_{\text{thres}}$ = 0.25, same below
                                        & 63.27 & 32.47 & 64.29 & 64.91 & 68.26 & 1  & 24 & 1045 & 1894  \\
(b) change to 2D Kalman Filter        & & 90.17 & 42.99 & 77.99 & 81.95 & 78.98 & 7  & 43 & 684 & 821 \\ 
(c) add angular velocity $v_{\theta}$ & & 93.29 & 45.44 & 77.40 & 86.16 & 78.39 & 0  & 16 & 365 & 795 \\
(d) remove orientation correction     & & 92.87 & 45.04 & 76.73 & 85.62 & 76.93 & 0  & 50 & 418 & 787 \\ 
(e) $\text{IoU}_\text{min}=0.1$       & & 92.43 & 45.25 & 77.44 & 85.63 & 78.47 & 0  & 18 & 366 & 838 \\ 
(f) $\text{IoU}_\text{min}=0.25$      & & 86.70 & 40.05 & 73.85 & 79.91 & 79.03 & 19 & 34 & 342 & 1322 \\ 
(g) $\text{Bir}_\text{min}=1$         & & 91.51 & 43.60 & 79.06 & 82.17 & 78.26 & 4  & 21 & 797 & 693 \\
(h) $\text{Bir}_\text{min}=5$         & & 90.56 & 43.49 & 75.46 & 84.89 & 78.69 & 0  & 13 & 278 & 988 \\
(i) $\text{Age}_\text{max}=1$         & & 90.89 & 43.60 & 75.86 & 83.96 & 78.90 & 0  & 43 & 380 & 964 \\ 
(j) $\text{Age}_\text{max}=3$         & & 91.26 & 44.48 & 77.17 & 84.75 & 78.21 & 0  & 13 & 503 & 775 \\
\midrule
(k) \textbf{Ours}                     & & 93.28 & 45.43 & 77.41 & 86.24 & 78.43 & 0  & 15 & 365 & 788 \\ 
\bottomrule
\end{tabular}}
\vspace{-0.3cm}
\label{tab:ablation}
\end{table*}

\begin{figure*}[t]
\begin{center}
\includegraphics[trim=0cm 6.6cm 0.1cm 0cm, clip=true, width=0.49\linewidth]{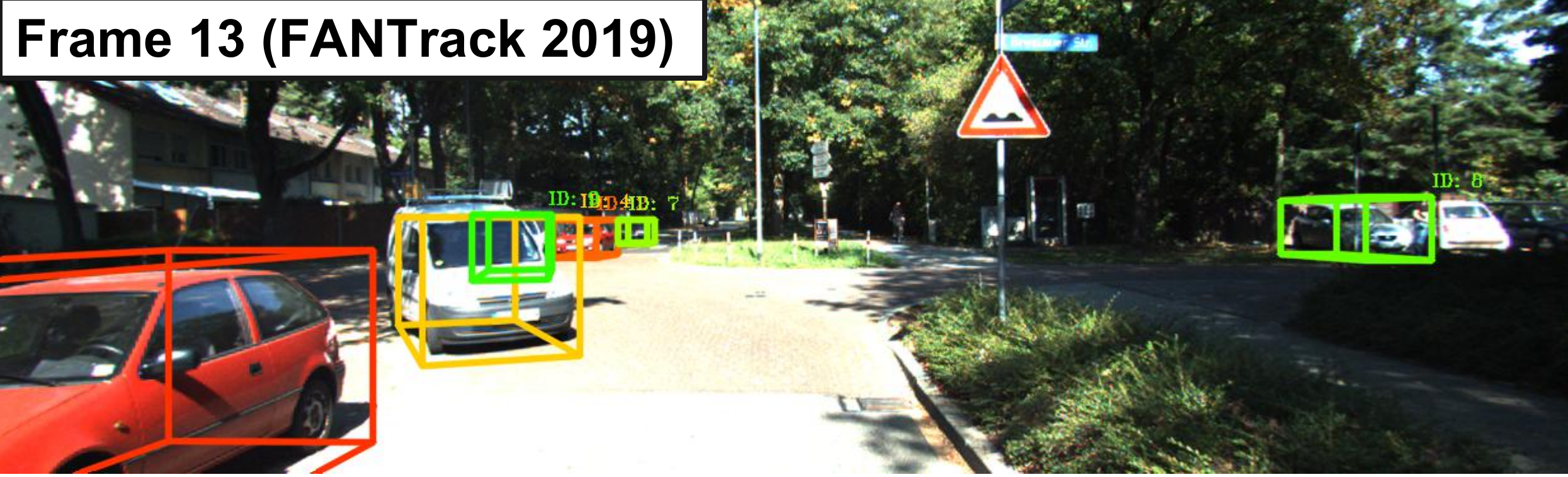}
\includegraphics[trim=0cm 6.6cm 0.1cm 0cm, clip=true, width=0.49\linewidth]{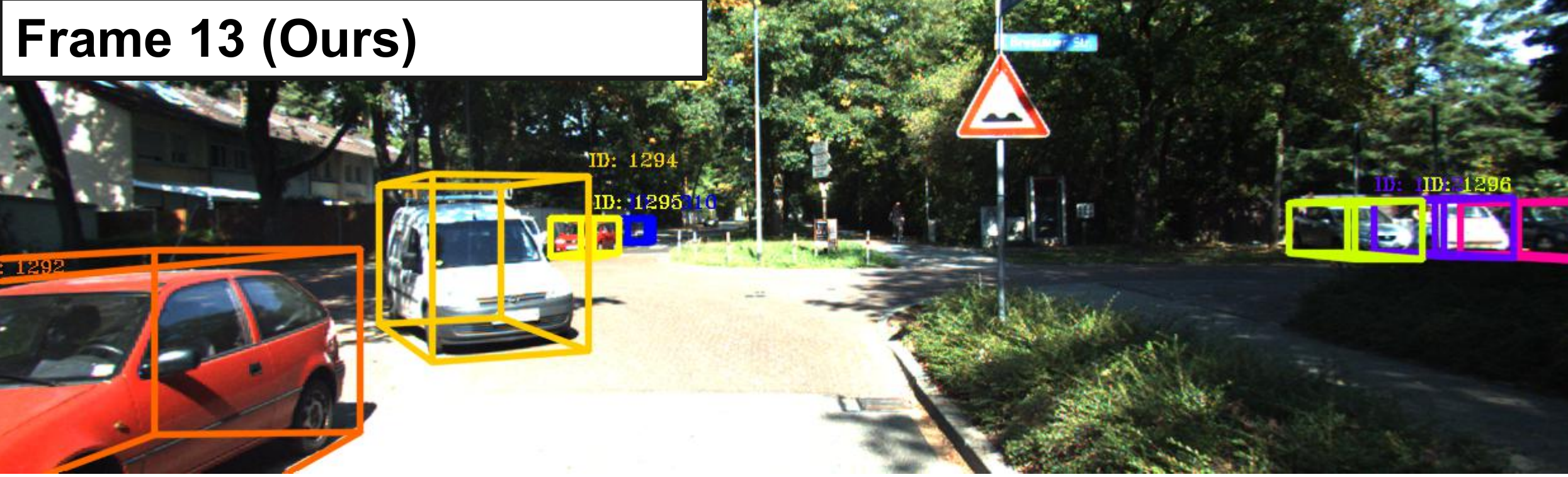}
\includegraphics[trim=0cm 6.6cm 0.1cm 0cm, clip=true, width=0.49\linewidth]{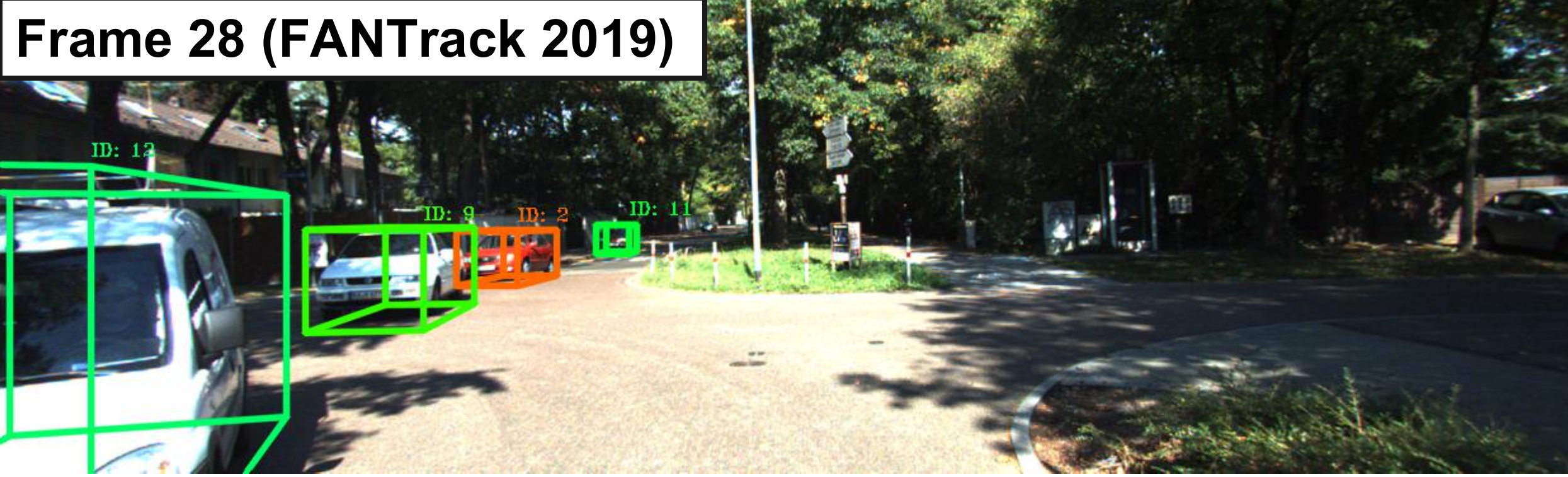}
\includegraphics[trim=0cm 6.6cm 0.1cm 0cm, clip=true, width=0.49\linewidth]{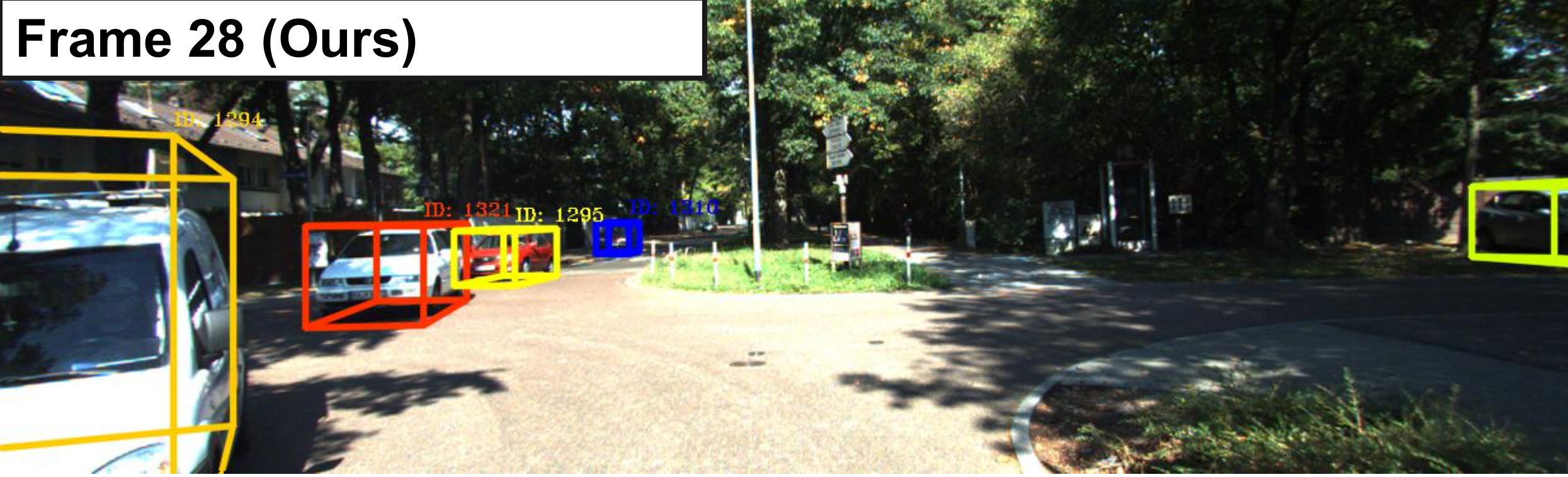}
\includegraphics[trim=0cm 6.6cm 0.1cm 0cm, clip=true, width=0.49\linewidth]{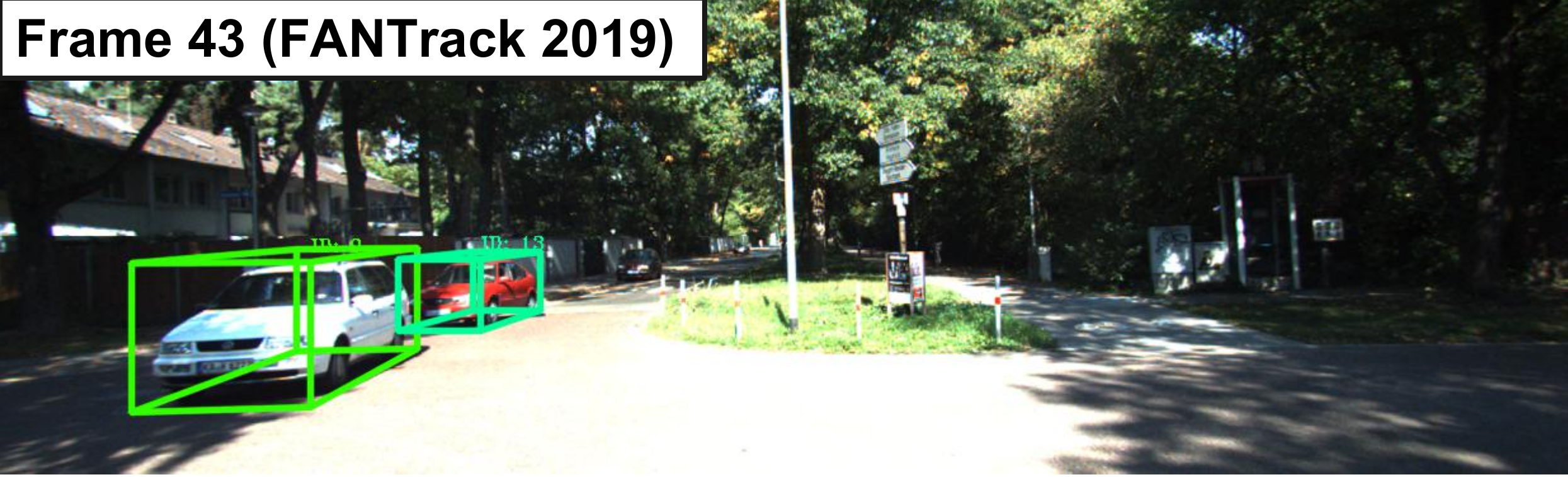}
\includegraphics[trim=0cm 6.6cm 0.1cm 0cm, clip=true, width=0.49\linewidth]{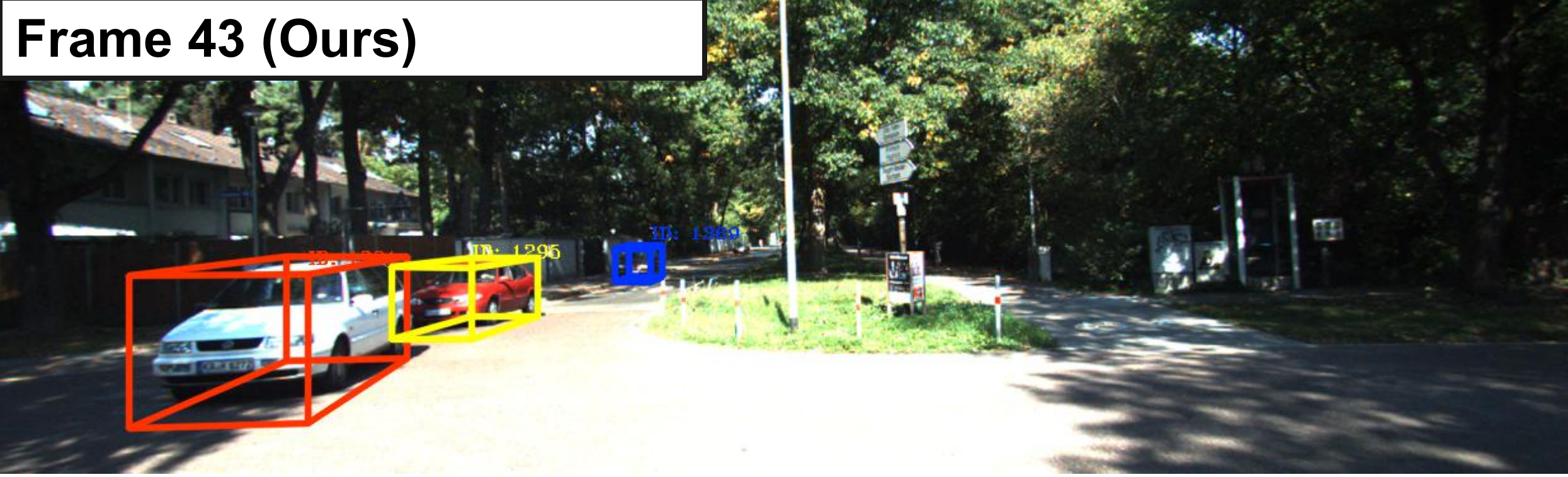}
\end{center}
\vspace{-0.55cm}
\caption{Qualitative comparison between FANTrack~\cite{Baser2019} (\textbf{left}) and our system (\textbf{right}) on the sequence 3 of the KITTI \textbf{test} set.} 
\label{fig:qua}
\vspace{-0.5cm}
\end{figure*}

\subsection{Experimental Results}

\noindent\textbf{Results for Cars on the KITTI val set}. We summarize the results in Table \ref{tab:3dcomparison}. Our proposed 3D MOT system consistently outperforms other modern 3D MOT systems in all metrics when using different matching criteria (\emph{e.g.}, 3D $\text{IoU}_{\text{thres}}$ = 0.25, 0.5, and 0.7). As a result, we establish new state-of-the-art 3D MOT performance on the KITTI val set for Cars and achieve an impressive zero identity switch.

\vspace{1.5mm}\noindent\textbf{Results for Pedestrians and Cyclists}. In addition to evaluate on cars, we also report our 3D MOT performance for other objects such as pedestrians and cyclists on the KITTI val set in Table \ref{tab:3dcomparison_pedcyc}. Although tracking of pedestrians and cyclists is more challenging than cars due to the small size of the objects, we show strong performance of our 3D MOT system.

\vspace{1.5mm}\noindent\textbf{Results for all objects on the nuScenes val set}. In addition to evaluate on the KITTI dataset, we also report 3D MOT results on the nuScenes val set in Table \ref{tab:3dcomparison_nuscene}. We emphasize that the nuScenes dataset is more challenging than KITTI due to sparse LiDAR point cloud inputs, complex scenes, and a low frame rate. Therefore, 3D detections on nuScenes are of significantly lower quality than 3D detections on KITTI, resulting in that all 3D MOT systems have a lower absolute performance on nuScenes. Our 3D MOT system still outperforms other 3D MOT systems in all metrics.

\vspace{1.5mm}\noindent\textbf{Inference Time.} We compare inference time of all methods in the last column of Table \ref{tab:3dcomparison}. Our 3D MOT system (excluding the 3D detector part) runs at a rate of $207.4$ FPS on the KITTI val set without the need of GPUs, achieving the fastest speed among other 3D MOT systems in Table \ref{tab:3dcomparison}.

\vspace{1.5mm}\noindent\textbf{Qualitative Comparison.} We show qualitative comparison between our 3D MOT system and \cite{Baser2019} and in Fig. \ref{fig:qua}. The 3D tracking results are visualized on the image with colored 3D bounding boxes where the color represents the object identity. We can see that the results of FANTrack (left) contain a few identity switches and miss tracking for objects at the rightmost of the image while our system (right) does not have these issues on the example sequence. We provide more qualitative results of our 3D MOT system in our demo video, which demonstrates that (1) our system, requiring no training, does not have the over-fitting issue on the dataset and (2) our system often produces more stable results and has fewer identity switches and jittered bounding boxes.


\vspace{-0.05cm}
\subsection{Ablation Study}

We conduct all ablative analysis for cars on the KITTI val set using the proposed 3D MOT evaluation tool along with the new metrics, which is summarized in Table \ref{tab:ablation}.

\vspace{1.5mm}\noindent\textbf{Effect of 3D Detection Quality.} In Table \ref{tab:ablation} (a), we switch the 3D detection module from \cite{Shi2019} to \cite{Weng2019}. The distinction lies in that \cite{Shi2019} requires a LiDAR point cloud as input while \cite{Weng2019} only requires a single image. As a result, the quality of 3D detections produced by the monocular 3D detector \cite{Weng2019} is much lower than the LiDAR-based 3D detector \cite{Shi2019} (see \cite{Shi2019, Weng2019} for details). We can see that the 3D MOT performance in (k) is also better than (a), suggesting that 3D detection quality is crucial to the performance of 3D MOT systems.

\vspace{1.5mm}\noindent\textbf{3D v.s. 2D Kalman Filter.} We replace the 3D Kalman filter in our final model (k) with a 2D Kalman filter \cite{Bewley2016} in (b). Specifically, we define the state space of an object trajectory $T$=$(x, y, a, r, s, v_x, v_y, v_a)$, where ($x$, $y$) is the object's 2D location, $a$ is the 2D box area, $r$ is the aspect ratio and ($v_x$, $v_y$, $v_a$) denote the velocity in the 2D image plane. We observed that using the 3D Kalman filter in (k) reduces the IDS from $7$ to $0$ and FRAG from $43$ to $15$, which we believe it is due to the fact that tracking in the 3D space can help resolve the depth ambiguity that exists if tracking in the 2D image plane. Overall, the absolute sAMOTA, AMOTA and MOTA values are improved by $3\%$ to $4\%$.

\vspace{1.5mm}\noindent\textbf{Effect of Angular Velocity $v_\theta$.} We add $v_\theta$ to the state space so that the state space of a trajectory $T$ = ($x$, $y$, $z$, $\theta$, $l$, $w$, $h$, $s$, $v_x$, $v_y$, $v_z$, $v_\theta$) in Table \ref{tab:ablation} (c). We observed that, compared to (k), adding $v_\theta$ improves sAMOTA and AMOTA by $0.01\%$ and decreases AMOTP and MOTA by up to $0.08\%$. This shows that adding the angular velocity or not does not really have a clear impact on the performance for all metrics. Therefore, we simply do not include the angular velocity in the state space of our final system for simplicity.

\vspace{1.5mm}\noindent\textbf{Effect of Orientation Correction.} As mentioned in Section \ref{sec:update}, we use an orientation correction technique in our final system in Table \ref{tab:ablation} (k). Here, we experiment a variant without using the orientation correction in Table \ref{tab:ablation} (d). We observed that the orientation correction helps improve the performance in all metrics, suggesting that this technique is useful to our proposed 3D MOT system.

\vspace{1.5mm}\noindent\textbf{Effect of Threshold $\text{IoU}_{\text{min}}$.} We change $\text{IoU}_{\text{min}}$=$0.01$ in (k) to $\text{IoU}_{\text{min}}$=$0.1$ in (e) and $\text{IoU}_{\text{min}}$=$0.25$ in (f). We observed that increasing $\text{IoU}_{\text{min}}$ leads to a consistent drop in all metrics.

\vspace{1.5mm}\noindent\textbf{Effect of $\text{Bir}_{\text{min}}$.} We adjust $\text{Bir}_{\text{min}}$=$3$ in (k) to $\text{Bir}_{\text{min}}$=$1$ in (g) and $\text{Bir}_{\text{min}}$=$5$ in (h). We show that using either $\text{Bir}_{\text{min}}$=$1$ (\emph{i.e.}, creating a new trajectory immediately for an unmatched detection) or $\text{Bir}_{\text{min}}$=$5$ (\emph{i.e.}, creating a new trajectory after an unmatched detection is matched in next five frames) leads to inferior performance in sAMOTA, AMOTP and MOTA, suggesting that using $\text{Bir}_{\text{min}}$=$3$ is the best. 

\vspace{1.5mm}\noindent\textbf{Effect of $\text{Age}_{\text{max}}$.} We verify the effect of $\text{Age}_{\text{max}}$ by decreasing it to $\text{Age}_{\text{max}}$=$1$ in (i) and increasing it to $\text{Age}_{\text{max}}$=$3$ in (j). We show that both (i) and (j) result in a drop in sAMOTA, AMOTA and MOTA, suggesting that $\text{Age}_{\text{max}}$=$2$ (\emph{i.e.} keep tracking the unmatched trajectories $T_{\text{unmatch}}$ in next two frames) in our final model (k) is the best choice.


\vspace{-0.1cm}
\section{Conclusion}

We proposed an accurate, simple and real-time system for online 3D MOT. Also, a new 3D MOT evaluation tool along with three new metrics was proposed to standardize future 3D MOT evaluation. Through extensive experiments on the KITTI and nuScenes 3D MOT datasets, our system establishes new state-of-the-art 3D MOT performance while achieving the fastest speed. We hope that our system will serve as a solid baseline on which others can easily build on to advance the state-of-the-art in 3D MOT. 





\vspace{-0.1cm}
\section*{ACKNOWLEDGMENT}

This work was funded in part by the Department of Homeland Security award 2017-DN-077-ER0001. Also, we thank the authors of SORT \cite{Bewley2016}, which inspired our work.

\bibliographystyle{IEEEtran}
\bibliography{IEEEabrv,main}


\end{document}